\documentclass[runningheads,a4paper]{llncs}

\usepackage{mathrsfs}
\usepackage{amssymb}
\usepackage{graphicx}
\usepackage{amsmath}

\newcommand{\linea}{\noindent\rule{1.0\textwidth}{1pt}}

\begin{document}

\mainmatter  

\title{Message passing for quantified Boolean formulas}

\author{Pan Zhang\inst{1} \and Abolfazl Ramezanpour\inst{1} \and Lenka Zdeborov\'a\inst{2} \and Riccardo Zecchina\inst{1} }

\institute{Politecnico di Torino, Corso Duca degli Abruzzi 24, I-10129 Torino, Italy \and Institut de Physique Theorique, IPhT, CEA Saclay,
and URA 2306, CNRS, 91191 Gif-sur-Yvette, France.}

\maketitle

\begin{abstract}
We introduce two types of message passing algorithms for quantified
  Boolean formulas (QBF). The first type is a message passing based heuristics that can prove
  unsatisfiability of the QBF by assigning the universal variables in
  such a way that the remaining formula is unsatisfiable. 
  In the second type, we use message passing to guide branching heuristics of a Davis-Putnam
Logemann-Loveland (DPLL) complete solver.  
  Numerical experiments show that on random QBFs our branching
  heuristics gives robust exponential efficiency gain with respect to
  the state-of-art solvers. We also manage to solve some previously
  unsolved benchmarks from the QBFLIB library. Apart from this our study sheds
  light on using message passing in small systems and as subroutines
  in complete solvers.
\end{abstract}

\section{introduction}
Satisfiability of quantified Boolean formulas (QBF) is a
generalization of the  Boolean satisfiability problem (SAT) where universal
quantifiers are added to the existential ones.  QBF
are useful in modeling practical problems harder than NP
e.g. planning, verification or combinatorial game playing.  Algorithmic
complexity of QBF ranges in the polynomial hierarchy up to PSPACE. 

Message passing algorithms are used in a wide range of algorithmically
hard problems, from
constraint satisfaction problems such as the satisfiability problem
\cite{MPZ02} to gene regulation network
reconstruction \cite{Weigt_etal_PNAS_2009}, error correcting codes \cite{EC_2000}, or
compressed sensing \cite{CS}.
Message passing algorithms can  be very efficient for large random systems where complete algorithms
cannot be applied.  However, on small systems sizes or for structured
SAT problems the use of message passing has been  so far very limited. For
instance there have been attempts to solve small satisfiability problems by using message passing in a
complete algorithm \cite{Hsu_etal_sat}.
However, according to these result it seems that message passing is helpful only
for satisfiable random formulas, where complete algorithms are less efficient than stochastic local search algorithms.

In this paper we introduce two types of message passing algorithms for
quantified Boolean Formulas, one as a heuristic scheme and
another as a subroutine for a complete solver.
We mainly consider QBF with two alternations $\forall X\exists
Y\Phi$ with $X$ corresponding to the set of universal variables, $Y$ corresponding to the set of existential variables, and
$\Phi$ being a CNF formula. Such QBF are said to be satisfiable if for all configurations of the universal variables,
there exists an assignment of the existential variables that satisfies the
formula. Examples for the case with more alternations are presented in Sec.~\ref{sec:extention}.

Solving a QBF is in general more difficult than solving a SAT formula.
In case of random SAT formulas, the state-of-art SAT solvers e.g.
kcnfs \cite{Dequen_Dubois_JAR_2006} or
march \cite{march_dl} can solve hard satisfiable $3$-SAT instances with
$600$ variables, while the state-of-art QBF solvers e.g. QuBE7.2
\cite{qube_2001} can solve hard random QBF with number of variables only up
to $60$. The intuitive reason for this difference is that a much
larger search space needs to be explored in QBF. Another reason is
that, whereas for SAT formulas there are several good decision
heuristics based on e.g. look-ahead~\cite{Li_Anbulagan_ACM_1997_366} or
backbone-search~\cite{Dubois_Dequen_2001}, for QBF efficient decision heuristics are missing. Here we introduce belief
propagation based decision heuristics that provides considerable speed
up to the state of the art QBF solvers. A similar attempt has been
done in \cite{Yin_etal_2011} where, however, the use of survey
propagation instead of belief propagation was unfortunate as we discuss briefly later. 

For the SAT problem, besides complete algorithms based on DPLL
\cite{DP_1960,DPL_1962} or resolution, there are also many efficient
heuristic algorithms \cite{Biere:2009:HSV:1550723}.
However up to our best knowledge, all algorithms proposed for QBF were complete
except one \cite{Interian_findingsmall}.
Here we introduce  a new message passing based heuristics for proving
unsatisfiability  that improves over  the algorithm of \cite{Interian_findingsmall}.

\section{Definitions and formulas}
In this section we remind the definition of random QBF formulas that
we use as benchmarks for our algorithms \cite{Gent_Walsh_AAAI_99,Chen_etal_2005}. We also remind the standard formulas
for belief propagation and survey propagation update rules \cite{MPZ02,Yedidia_etal_TR2001-22,Braunstein_Mezard_Zecchina_RSA_2005}.

To express QBF with $t$ alternations, we use $Q_1V_1Q_2V_2...Q_tV_t\Phi$ where $Q_n$ denotes quantifier $\exists$ or $\forall$ at $n$th 
alternation, $V_n$ denotes the set of variables at $n$th alternation and
$\Phi$ denotes the set of clauses. We evaluate satisfiability of QBF
in the following way. If one is able to find an assignment of the
universal variables for which no solution exists, the QBF is said
unsatisfiable, otherwise it is satisfiable. For instance, $\forall X
\exists Y\Phi$ denotes QBF with two alternations, and the set of universal $X$ and existential $Y$ variables. 
This formula is satisfiable if for every assignments of $X$, one can find an assignment for $Y$ such that $\Phi$ is satisfied. 
We will use the notation $N_u\equiv |X|$, $N_e\equiv |Y|$, $M$ for the number of clauses in $\Phi$, $\alpha_e\equiv M/N_e$ and $\alpha_u\equiv M/N_u$.

Several models for random QBF were proposed. In this paper, we consider the $(L,K)$ model
\cite{Chen_etal_2005}, and the model-B \cite{Gent_Walsh_AAAI_99}. In
model-B, each clause in $\Phi$ has $U$ 
universal variables and $V>0$ existential variables that are
selected randomly from the whole set of universal (resp. existential) variables.
The $(L,K)$ model is 
a special case of model-B formula with 2 alternations, which specifies a formula $\forall X\exists Y\Phi$ 
where each clause in $\Phi$ contains $L+K$ variables, $L$ from $X$ and $K$ from $Y$.

In message passing algorithms, belief
propagation \cite{Yedidia_etal_TR2001-22} and survey
propagation \cite{MPZ02},  we define $\{\psi_i^s\}$ to be the marginal
probability that variable $i$ takes assignment $s$ among all the
solutions (BP) or among all the solution clusters (SP). In BP $s$ has two
possible values $+$ or $-$, with  $\psi_i^+ + \psi_i^-=1$, in SP $s$
has three possible values $+$,$*$, and $-$ with $\psi_i^+ +\psi_i^*+
\psi_i^-=1$.
We say that a variable is biased if $\psi_i^+ \neq \psi_i^-$, the
larger the difference the larger bias the variable has. If
$\psi_i^+>\psi_i^-$ we define the bias of $i$ to be $\psi_i^+$, and if
$\psi_i^+<\psi_i^-$ we define the bias of $i$ to be $\psi_i^-$. 

Let us define $\partial_{+} i$ as the set of clauses to which $i$ belongs
non-negated, and $\partial_{-} i$ as the set of clauses to which $i$
belongs negated. Then the set of clauses to which variable
$i$ belongs can be written as $\partial i = \{a\} \cup {\cal S}_{ia} \cup
{\cal U}_{ia}$ where (a) if $i$ is not negated in $a$ then $ {\cal
  S}_{ia}=\partial_{+} i \setminus a$, ${\cal U}_{ia}=\partial_{-} i$, and
(b) if $i$ is negated in $a$ then $ {\cal S}_{ia}=\partial_{-} i \setminus a$, ${\cal U}_{ia}=\partial_{+} i$.

The BP marginals are computed as 
\begin{equation}
 \psi^+_{i}=\frac{  \prod_{b\in\partial_{-} i} u_{b\to i} \prod_{b\in \partial_{+} i} (1-u_{b\to i})
 }{\prod_{b\in \partial_{+} i} u_{b\to i}   \prod_{b\in \partial_{-} i}
     (1-u_{b\to i})  +\prod_{b\in \partial_{-} i} u_{b\to i}
     \prod_{b\in \partial_{+} i} (1-u_{b\to i})   }\, ,
\end{equation} 
where messages $u_{b\to i}$ are a fixed point of the following iterative equations 
\begin{eqnarray}\label{eq:bp}
    \psi_{i\to a}&=&\frac{\prod_{b\in{\cal S}_{ia}} u_{b\to i}
      \prod_{b\in{\cal U}_{ia}} (1-u_{b\to i})
    }{\prod_{b\in{\cal S}_{ia}} u_{b\to i}   \prod_{b\in{\cal U}_{ia}}
      (1-u_{b\to i})  +\prod_{b\in{\cal U}_{ia}} u_{b\to i}
      \prod_{b\in{\cal S}_{ia}} (1-u_{b\to i})  }\, ,\nonumber\\
    u_{b\to i}&=&\frac{ 1-\prod_{j\in\partial b\backslash i}\psi_{j\to
        b} }{ 2-\prod_{j\in\partial b\backslash i}\psi_{j\to
        b}}\, .
\end{eqnarray} 
 Iteration equations of SP are written as:
\begin{eqnarray}
\psi^U_{i\to a}&=& \frac{1}{C_{i\to a}}\left[1-\prod_{ b\in
  {\cal U}_{ia}}u_{b\to i}\right] \prod_{ b\in {\cal S}_{ia} 
 }u_{b\to i} \, ,\nonumber \\
  \psi^S_{i\to a}&=&\frac{1}{C_{i\to a}}
 \left[1-\prod_{ b\in {\cal S}_{ia}}u_{b\to i}\right]
 \prod_{ b\in  {\cal U}_{ia} }u_{b\to i}\, ,\nonumber\\ 
 \psi^*_{i\to a}&=& \frac{1}{C_{i\to a}}\prod_{ b\in \partial
   i\setminus a }u_{b\to i}\, ,\nonumber\\
 u_{a \to i}&=&1-\prod_{j \in \partial a\backslash i}\psi_{j\to a}^{U},
\label{eq:speta1}
\end{eqnarray}
where $C_{i\to a}$ is a normalization constant ensuring $\psi^+_{i\to
  a}+\psi^*_{i\to a}+\psi^-_{i\to a}=1$. SP marginals are computed by
\begin{eqnarray}
\psi^+_{i}&=& \frac{1}{C_{i}}\left[1-\prod_{ b\in
 \partial_{+} i}u_{b\to i}\right] \prod_{ b\in \partial_{-} i 
 }u_{b\to i}\, , \nonumber \\
  \psi^-_{i}&=&\frac{1}{C_{i}}
 \left[1-\prod_{ b\in \partial_{-} i}u_{b\to i}\right]
 \prod_{ b\in \partial_{+} i  }u_{b\to i}\, ,\nonumber\\
 \psi^*_{i}&=& \frac{1}{C_{i}}\prod_{ b\in \partial i}u_{b\to i}\, ,\nonumber\\
\label{eq:spmar}
\end{eqnarray}
where $C_{i}$ is again a normalization constant.

\section{Heuristic algorithm for proving unsatisfiability}

Proving unsatisfiability for the two-level QBF can be done by finding
an assignment of the universal variables that leaves the existential part
of the formula unsatisfiable. One strategy is
to find a configuration of the universal variables
that leaves the largest possible number of clauses unsatisfied; see e.g. WalkMinQBF \cite{Interian_findingsmall}.
This, however, leads to heuristic algorithms that do not use in any way the
existential part of the QBF. Here we suggest and test a belief
propagation decimation heuristics
for proving unsatisfiability that takes into account the whole formula
and outperforms significantly the previously known heuristics.

Our belief propagation decimation heuristics  (BPDU) for proving
unsatisfiability of QBF works as follows: Input is
the QBF formula $\forall X\exists Y\Phi$. We
disregard for a moment the quantifiers in the QBF formula and run randomly
initialized BP on the whole formula (i.e. with both universal and
existential variables) till convergence or till the maximum number of
allowed iterations $t_{\rm max}$ is achieved (typically we use $t_{max}=300$ in our algorithms). We select the most biased
universal variable and fix it against the direction of the bias.  We
repeat the above steps until all the universal variables are assigned.  
Then we run a complete SAT
solver (in our case kcnfs \cite{Dequen_Dubois_JAR_2006,Dubois_Dequen_2001}) on the formula consisting of the existential variables and
clauses that were not satisfied by any of the universal variables. If
this remaining formula is unsatisfiable then we proved
unsatisfiability of the QBF. If the remaining formula is satisfiable
then the algorithm outputs "unknown", since there might be another
configuration of the universal variables that would leave the
remaining formula unsatisfiable.

A variation on the above heuristic algorithm is to run survey
propagation instead of belief propagation whenever survey propagation
converges to a nontrivial fixed point. In what follows we call this variation BPSPDU.

\begin{figure}[!ht]
\includegraphics[width=0.5\textwidth]{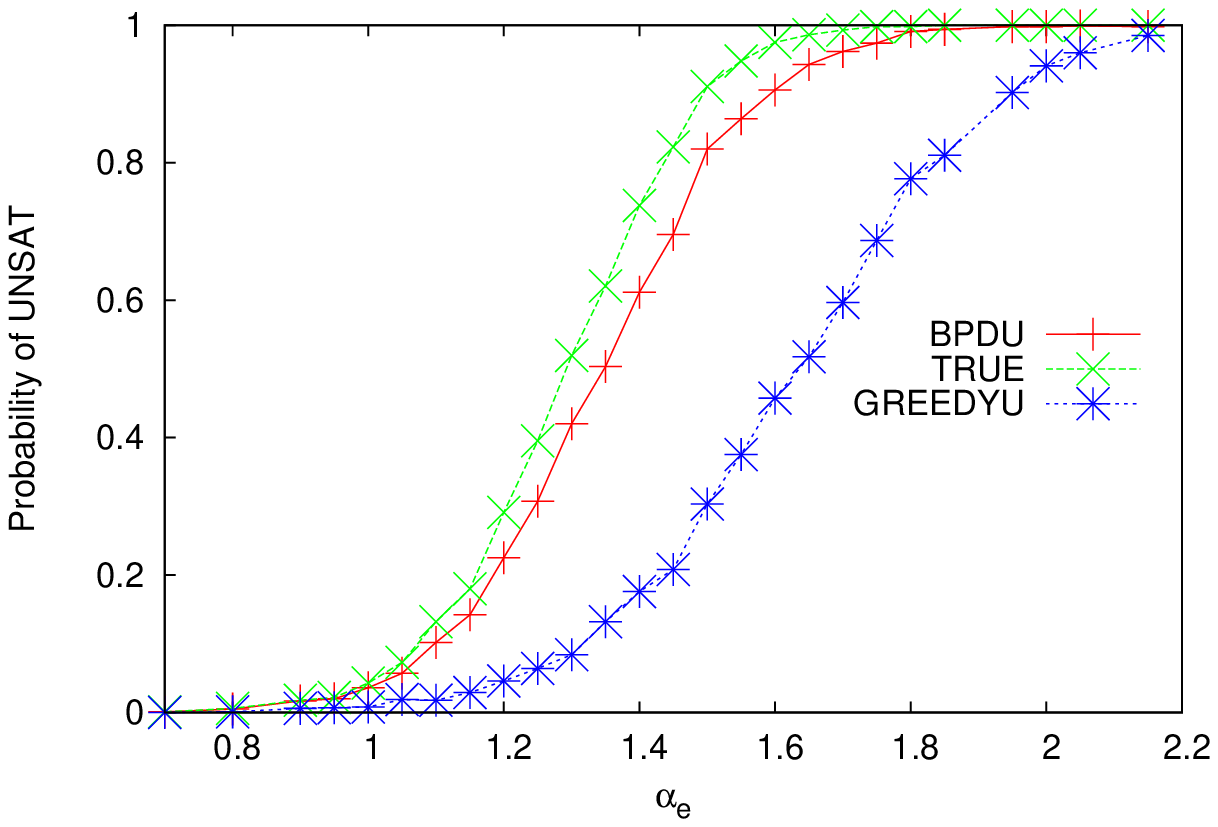}
\includegraphics[width=0.5\textwidth]{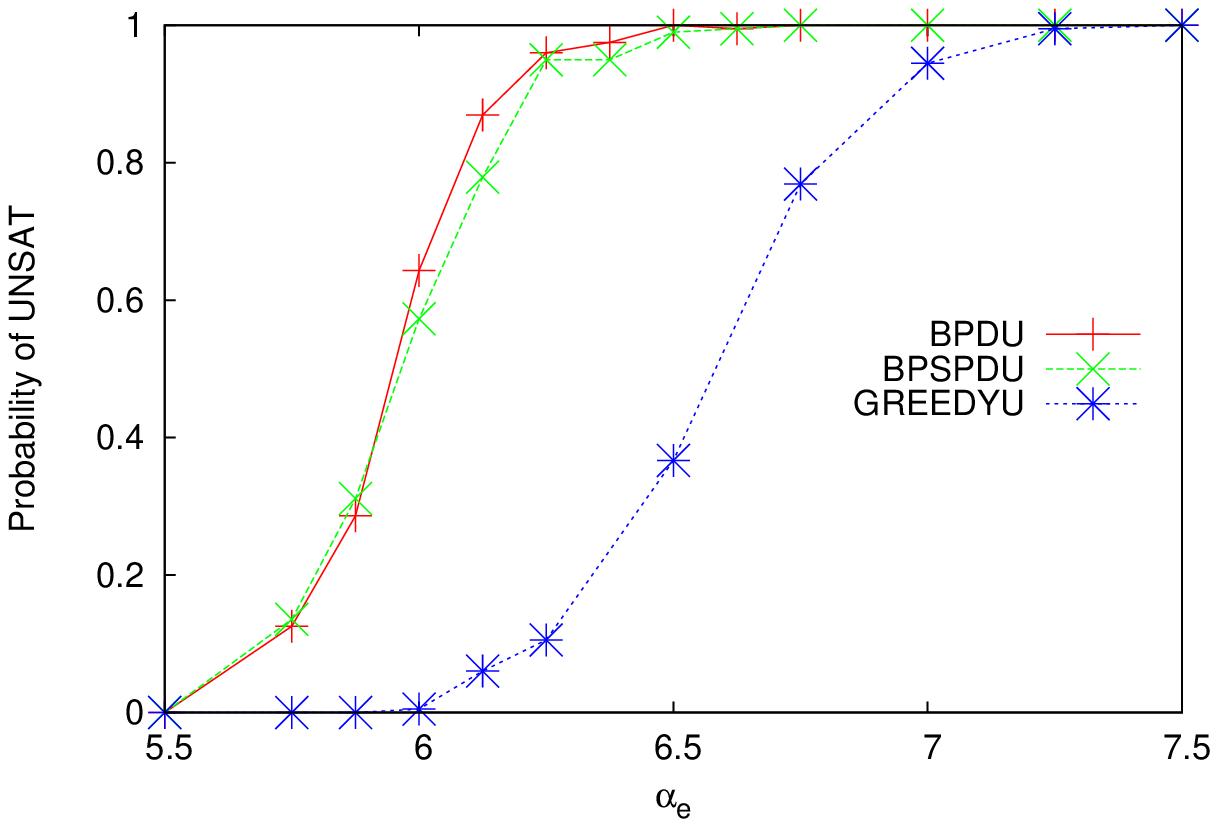}
\caption{ \label{fig:2_13_200}
Ratio of unsatisfiable formulas found by the various algorithms
discussed in the text for random $(1,2)$ QBF instances (left) and
$(1,3)$ QBF instances (right) with $N_u=N_e=200$ as a function of
$\alpha_e=M/N_e$.
}
\end{figure}

To evaluate the performance of the BPDU and BPSPDU heuristics we first
apply them on random $(1,2)$ and $(1,3)$ QBF instances with $N_e=N_u=200$
variables and with varying number of clauses (more clauses make the
formulas less likely to be satisfiable), see Fig.~\ref{fig:2_13_200}. 
For the $(1,2)$ case, a complete solver (e.g. QuBe7) can solve every
formula hence we can compare the fraction of unsatisfiable formulas
found by BPDU to the true fraction. For the $(1,3)$ case the size is
prohibitive for the use of a complete QBF solver. In both cases we also show
the fraction of unsatisfiable formulas found by the "greedy" strategy,
which fixes the universal variables in order to let the largest
possible number of unsatisfied clauses. In case of $(1,K)$ QBF this is
easy as every universal variable just needs to be set positive if it
appears negated in more clauses than non-negated,
and vice versa.  In Fig.~\ref{fig:2_13_200}  we see that the BPDU and
BPSPDU heuristics perform much better than the greedy strategy. And in
case of $(1,2)$ QBF the performance is not too far from the optimal.

\begin{figure}[!ht]
  \includegraphics[width=0.5\textwidth]{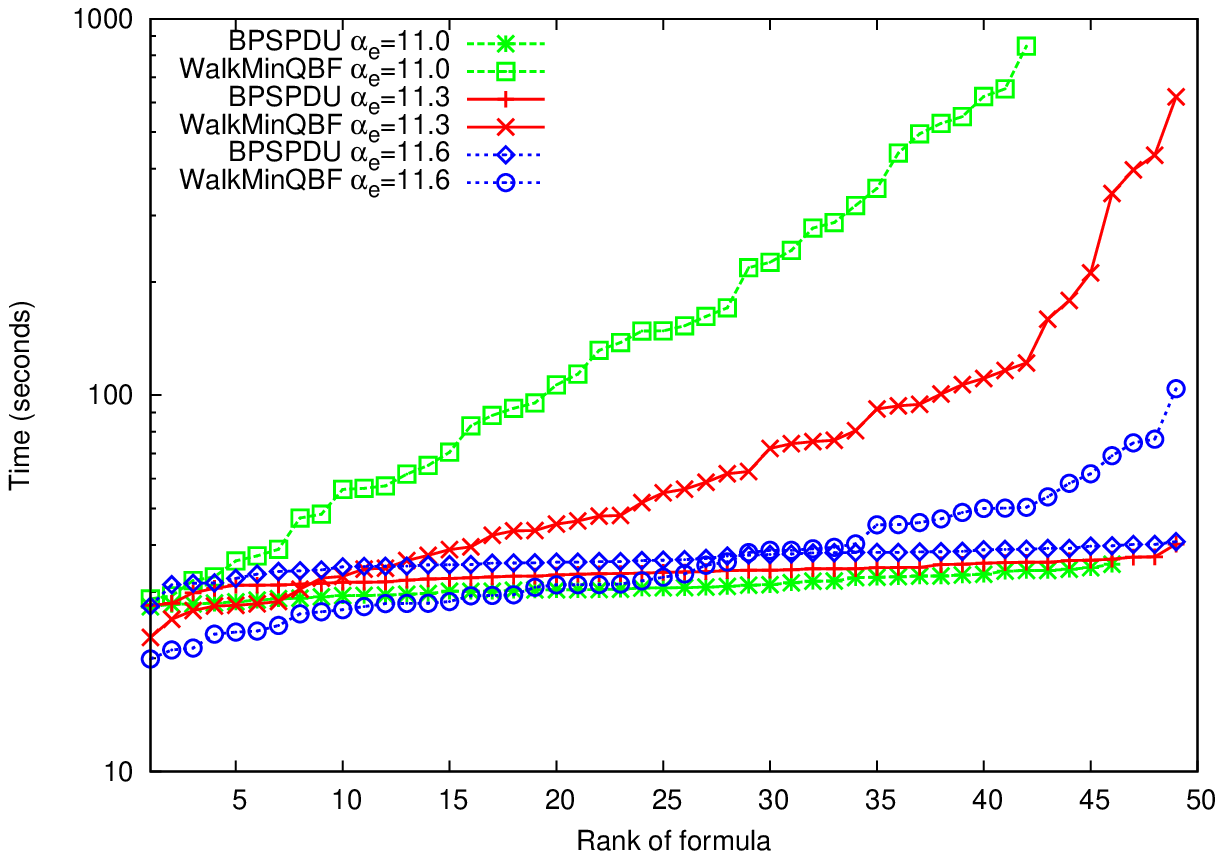}
  \includegraphics[width=0.5\textwidth]{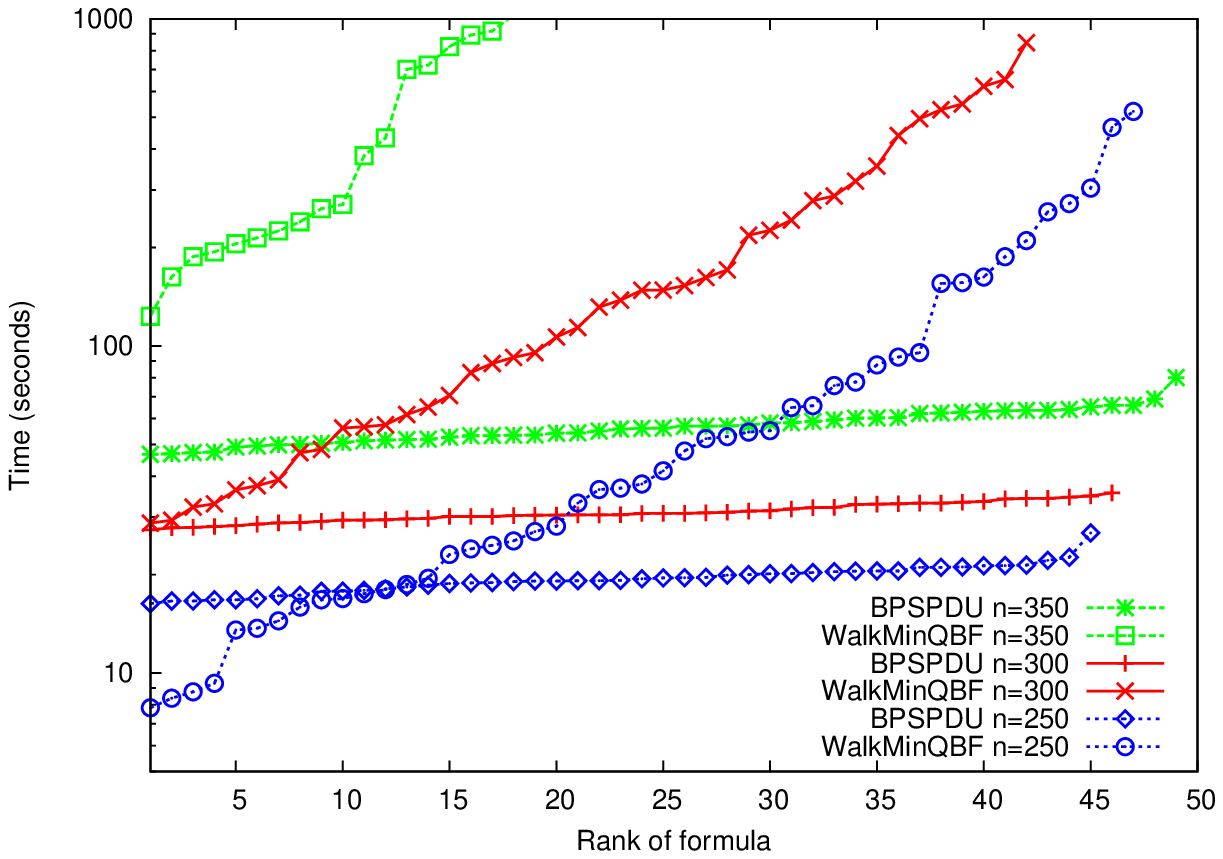}
\caption{ \label{fig:2_23}
Running time of the BPSPDU heuristics compared to WalkMinQBF on $50$ random
$(2,3)$ QBF benchmarks with the same number of variables $N_e=N_u=300$ but different $\alpha_e$ values (left), 
and the same $\alpha_e=11.0$ but different number of variable $n=N_e=N_u$ (right)}.
\end{figure}

In Fig.~\ref{fig:2_23} we compare the performance of the BPSPDU and
WalkminQBF heuristics for random $(2,3)$ instances. The WalkMinQBF aims
at setting the universal variables in order to maximize the number of
unsatisfied clauses, and then evaluates the remaining SAT formula with a complete SAT solver.
Note that the complete SAT
solver used by WalkminQBF and BPSPDU is the same,
kcnfs \cite{Dequen_Dubois_JAR_2006,Dubois_Dequen_2001}, so the
difference in performance comes only from the quality of the
universal configuration given by the two heuristics.


As the figure shows, formulas with larger number of clauses are easier for WalkMinQBF,
and the running time of the algorithm has larger fluctuations compared to that of BPSPDU.   
The difference between the two algorithms is clearer for larger problem instances;
with $N_e=N_u=350$, BPSPDU solves $49$ out of the $50$ instances, and WalkMinQBF solves only $17$ of them within $1000$ seconds.
The above results indicate that the universal configuration
suggested by BPSPDU is much better than the one suggested by WalkMinQBF.

\section{Message passing to guide QBF complete solvers}

When the BPDU or the BPSPDU algorithms introduced in the last section
output "unknown", there might be another configuration of the
universal variables that makes the formula unsatisfiable. Given that we
were fixing the universal variables starting with the most biased one,
it might be a good strategy to backtrack on the variables fixed in the
later stages. In this section we extend this idea into a complete DPLL-style solver, which is
using message passing to decide on which variables to branch next.

DPLL-style algorithms are the most efficient complete solvers for SAT
and QBF, they search the whole configurational space by backtracking.
The difference between DPLL for SAT and for QBF is that in QBF DPLL does
backtracking on existential variable when it encounters a contradiction, and
does backtracking on universal variable when it encounters a solution,
see e.g. \cite{Zhang_Malik_QBF_2006} for details. Besides the basic DPLL backtracking procedure, there are several
components in modern SAT and QBF solvers that lead to exponential
speed up, among the important ones are decision heuristics, unit-clause propagation,
non-chronological back-jumping, conflict and solution driven clause
learning \cite{Zhang_Malik_QBF_2006,Zhang_etal_leaning_2001}.
Our contribution concerns the \textit{decision heuristics} which is
used in order to decide which variable will be used in the next branch and which sign of the variable should be checked
first. Decision heuristics guides DPLL to the more relevant branches and
keeps it away from irrelevant branches.

Here we propose a decision heuristics that uses information coming from the result of belief
propagation (that was iterated till convergence or for $t_{\rm max}$ steps on the whole formula ignoring the
quantifiers). We propose to start branching with the more biased universal variables and
assign them first the less probable values. For the
existential variables we start branching also on the more biased
ones, but assign them the more probable values. The motivation is that this
will speed up the search of a universal configuration that will
leave the existential part of the QBF unsatisfiable, and the search
for a solution on the existential part.  In particular we propose two
decision heuristics.

In the BPH decision heuristics we simply order the variables according to their
bias, starting with the most biased one, and assign them first the
value opposite to the bias for the universal variables and according
to the bias for the existential ones.

In the BPDH decision heuristics we run BP and choose the most biased
unassigned variables which belong to the highest quantifier order. If this
variable is universal we assign it the value opposite to its bias, if
this variable is existential we assign it according to its bias. We
repeat BP on the simplified formula. Finally the BPDH heuristics will
branch variables in the same order as they were encountered in this
procedure. Most of the good decision heuristics for SAT and QBF solvers are dynamic,
which means that the branching sequence is being updated during the
run. Our BPH and BPDH decision heuristics are computationally heavier than
the other efficient decision heuristics e.g. VSIDS, MOMs or
failed-literal-detection \cite{Li_Anbulagan_ACM_1997_366}, so we use it only once at the very
beginning of the DPLL run.

We report performance of DPLL with our BPH and BPDH decision
heuristics on two levels. First level is using BPH and BPDH in the
pure DPLL (no features such as conflict and solution driving
back-jumping and clause learning included). The second level is using BPH in a state-of-art QBF solver QuBE7.2, which
is one of the fastest known solvers today.

\begin{figure}[!ht]
\includegraphics[width=0.5\textwidth]{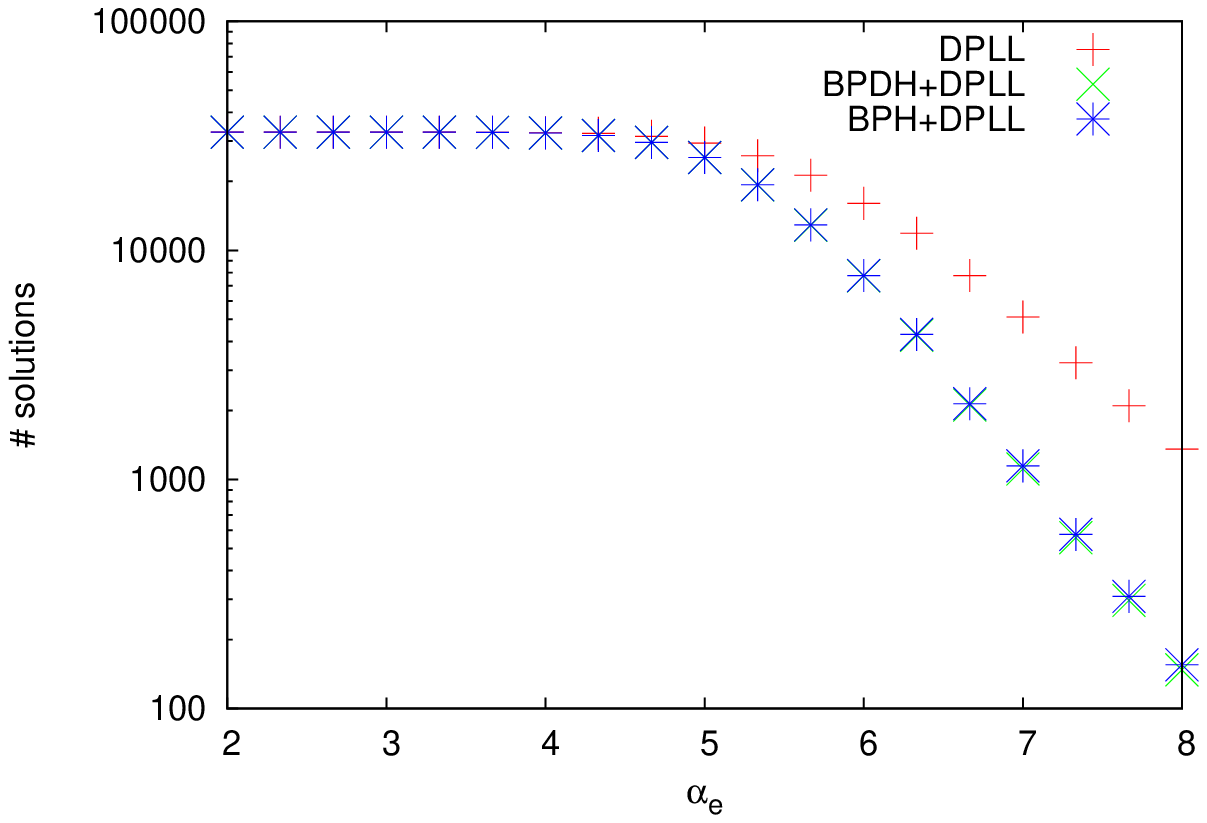}
\includegraphics[width=0.5\textwidth]{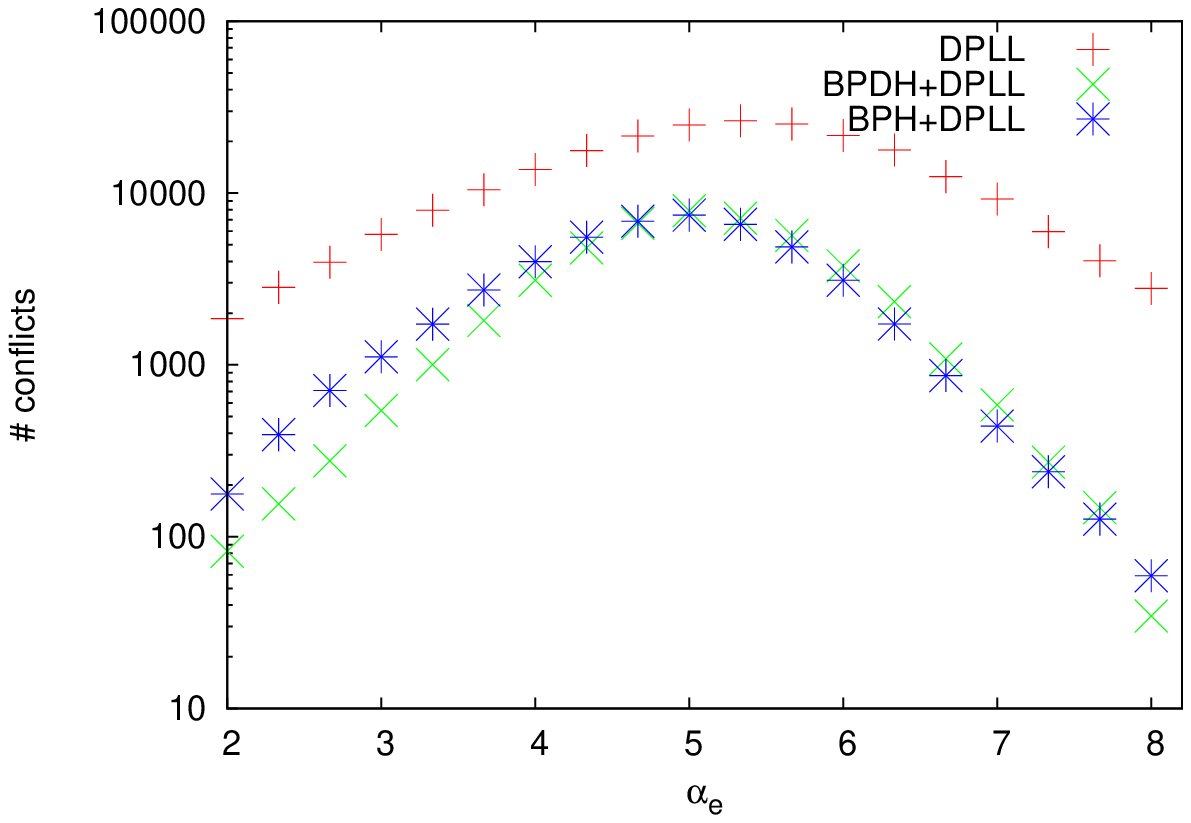}
\includegraphics[width=0.5\textwidth]{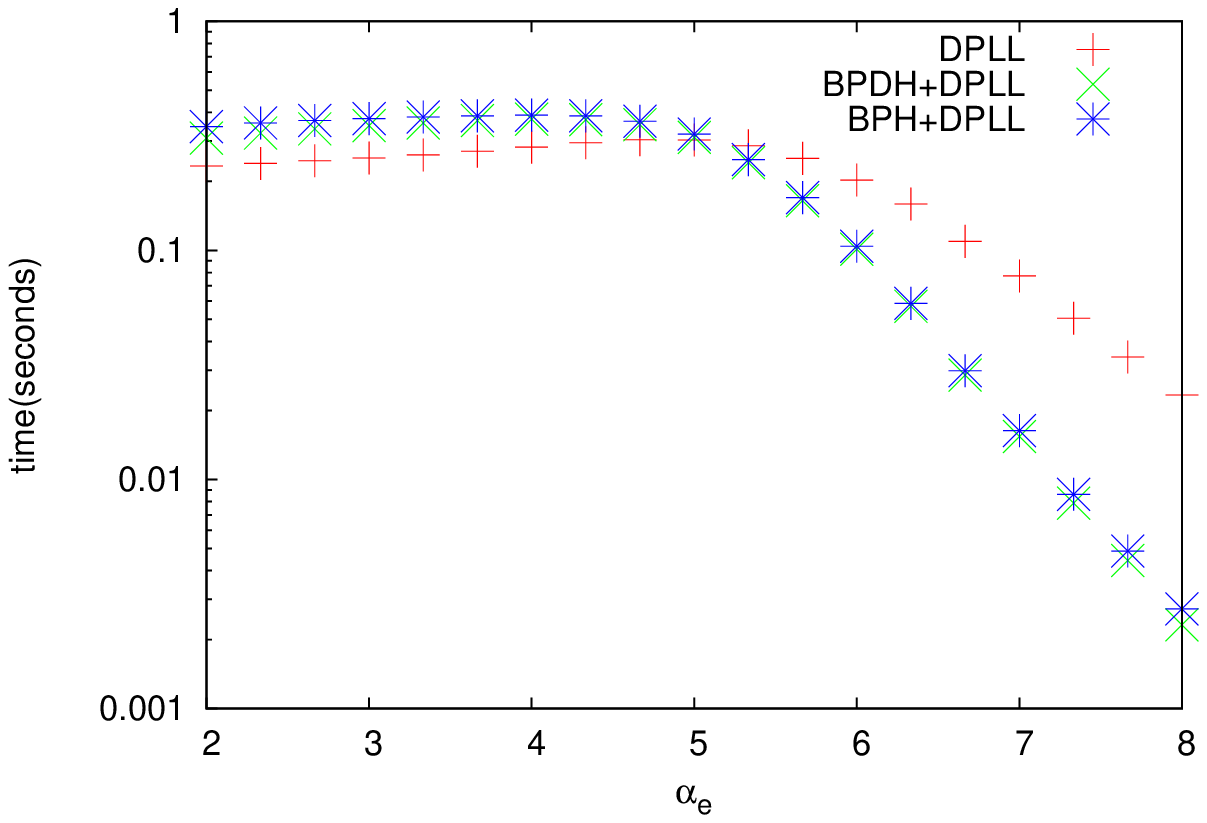}
\includegraphics[width=0.5\textwidth]{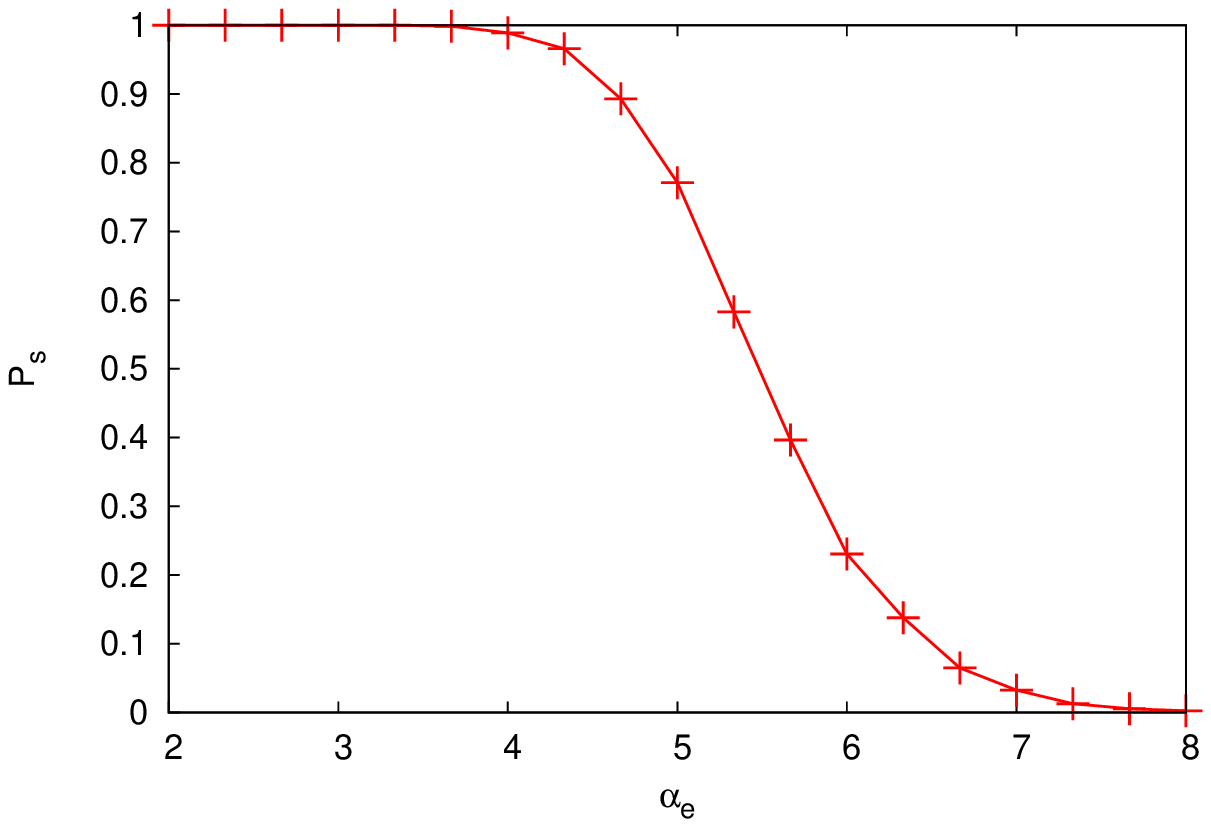}
\caption{ \label{fig:15con}
Number of solutions, number of conflicts, running time, and ratio of
satisfiable formulas $P_s$, as resulting from pure DPLL for solving random $(1,3)$ formulas, data are averaged over $10000$ instances.
$N_u=15$ and $N_e=15$.
}
\end{figure}

In Fig.~\ref{fig:15con} we plot the number of solutions and the number of
conflicts encountered in DPLL using a pure DPLL algorithm and those
encountered in DPLL with BPDH and BPH decision heuristics on random
$(1,3)$ QBF formulas. By pure DPLL we mean with no back-jumping nor clause learning, and the default decision heuristics is
VSIDS (Variable State Independent Decaying Sum) \cite{VSIDS,Zhang_etal_leaning_2001}, which is based on dynamic statistics of literal count as a score to order
literals (in case of pure DPLL, no learned clause contributes to the literal count).
Since pure DPLL is very CPU-time demanding, we use formulas with only $N_u=N_e=15$ variables.
We also plot the ratio of satisfiable
formulas $P_s$, and we can see that when almost all formulas are satisfiable,
the average
number of solutions encountered in DPLL is always $2^{15}$, because to
prove the satisfiability of a formula, pure DPLL has to scan all the $2^{N_u}$ universal configurations. When $P_s$
becomes smaller than one, DPLL with BPDH and BPH encounters much
smaller number of solutions than the pure DPLL. Fig.~\ref{fig:15con}
shows that in the whole range of parameters the
number of conflicts encountered by DPLL with BPDH and BPH is always
much smaller than DPLL with VSIDS decision heuristics.
The fewer solutions and conflicts encountered, the smaller search tree
is explored by the algorithm.

A better way to extract the full power of BPH and BPDH in DPLL-based
search is to use solution and conflict driven back-jumping and clause
learning. In clause learning, reasons of solutions and conflicts are
analyzed and stored as learned clauses in order to implement
non-chronological back-jumping to more relevant branches of the search
tree and to avoid the encounter of the same solutions or conflicts in the future search. With clause learning, DPLL does not have to
 explore $2^{N_u}$ satisfiable leaves of the search tree, and good
 decision heuristics could lead to a smaller number of solutions.
 We applied clause learning to the pure DPLL with and without
 BPH. Results show that with BPH and BPDH, both search tree size and
 running time are exponentially smaller than without BPH for both
 unsatisfiable  and satisfiable formulas.

As a next step we implemented message passing decision heuristics in
a state-of-art solver, we chose QuBE7.2, which uses solution and conflict driven clause learning, as the fastest known solver
today, and replaced the decision heuristics in QuBE7.2 by BPH.
We have also tried using BPH in other state-of-art solvers, and found 
quantitatively similar results.

\begin{figure}[!ht]
\includegraphics[width=0.5\textwidth]{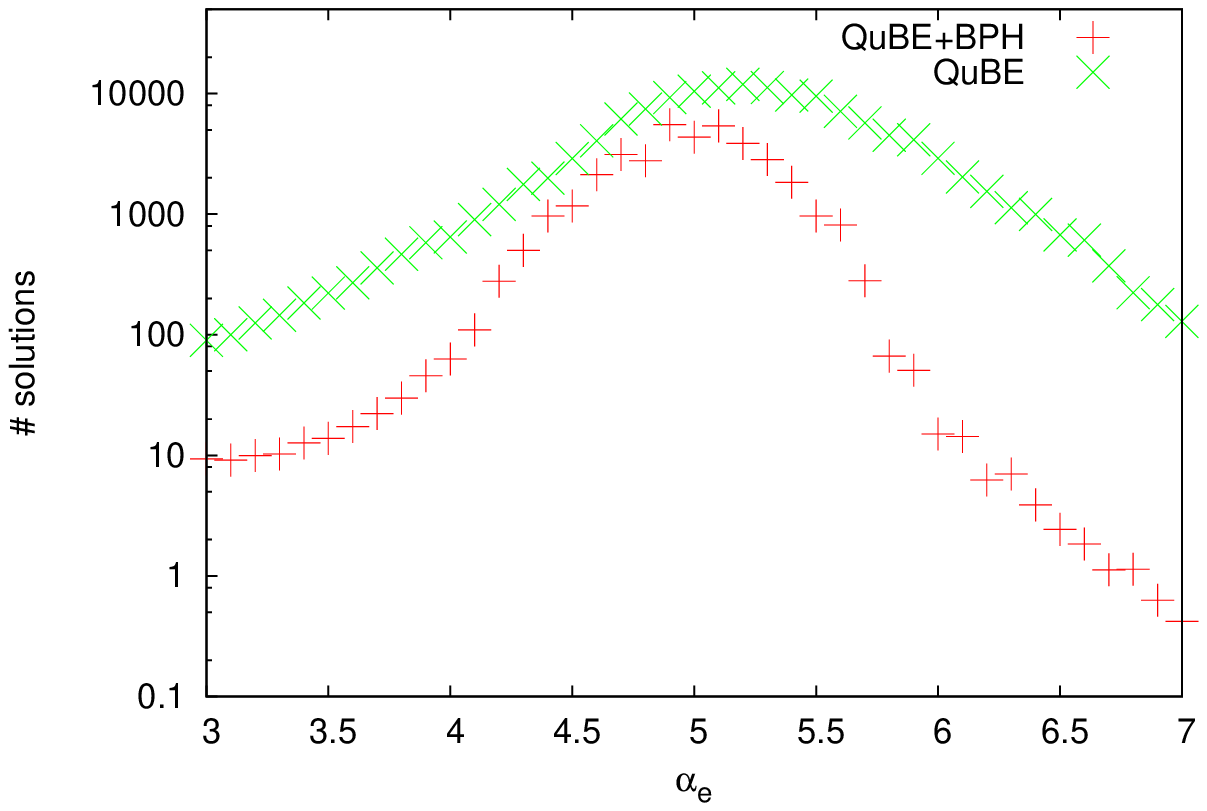}
\includegraphics[width=0.5\textwidth]{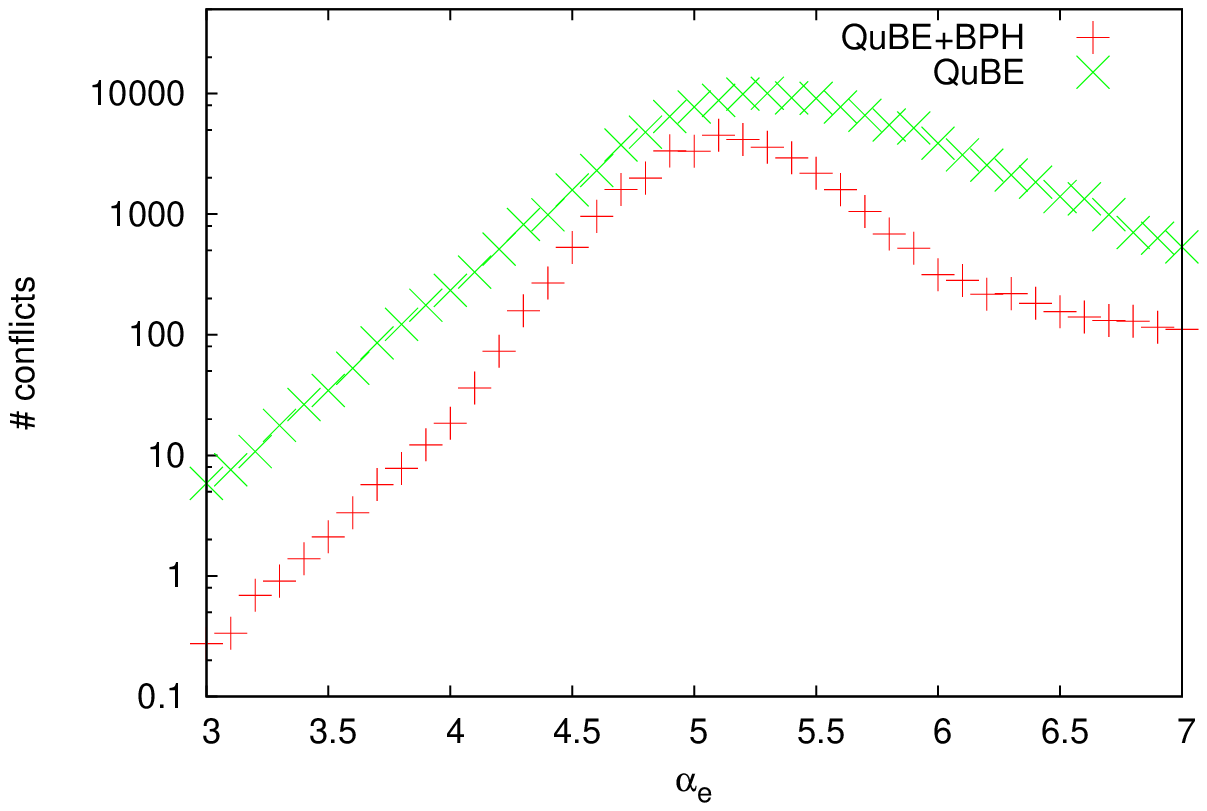}
\includegraphics[width=0.5\textwidth]{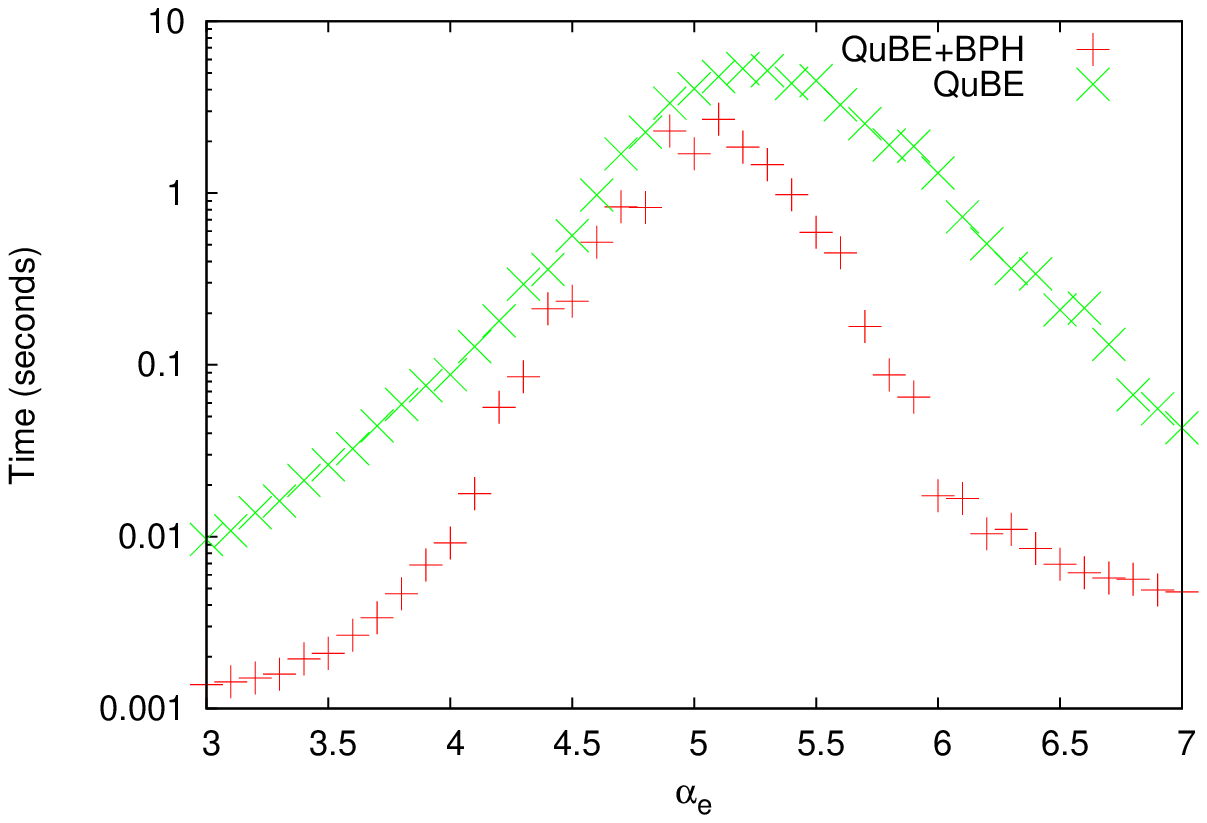}
\includegraphics[width=0.5\textwidth]{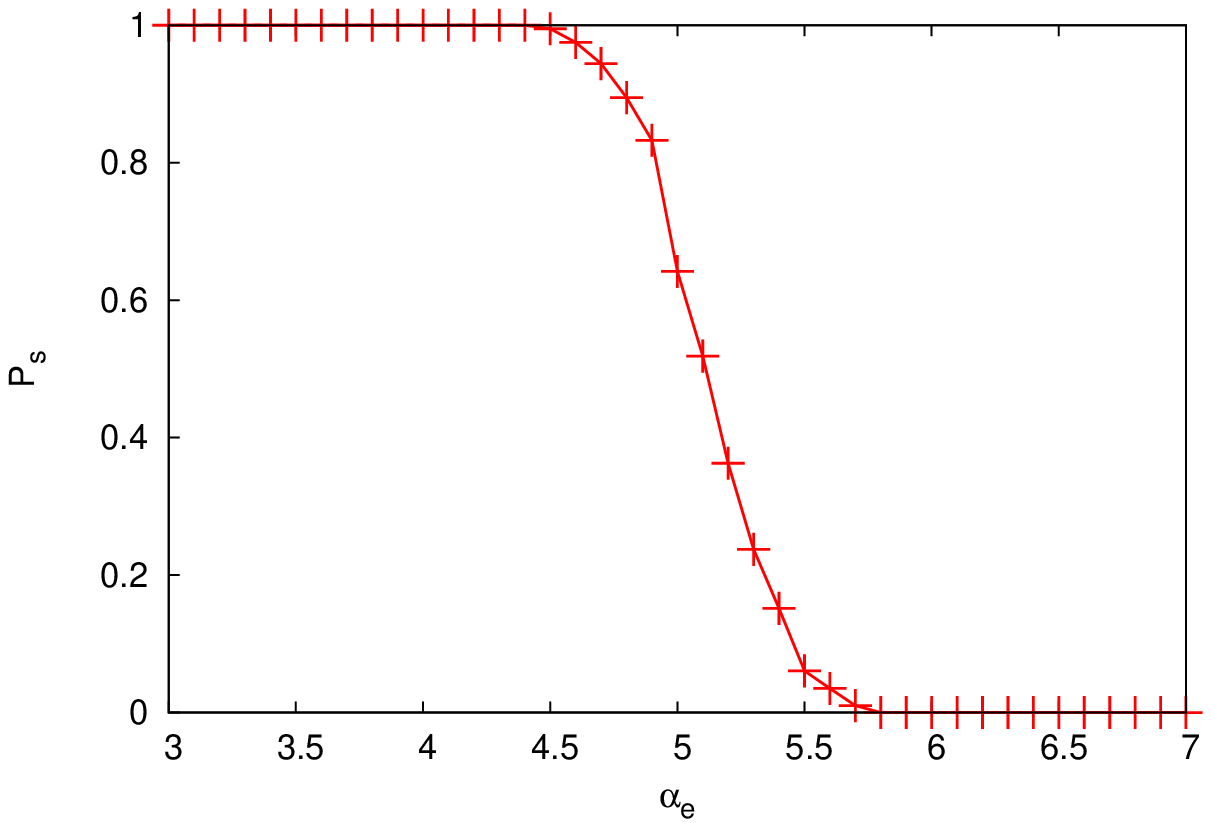}
\caption{ \label{fig:50con}
Number of conflicts, solutions and time used by DPLL of QuBE7.2 and QuBE7.2+BPH in solving random $(1,3)$ formulas, data are averaged over $200$ instances.
$N_u=50$ and $N_e=50$.
}
\end{figure}

\begin{figure}[!ht]
  \includegraphics[width=0.8\textwidth]{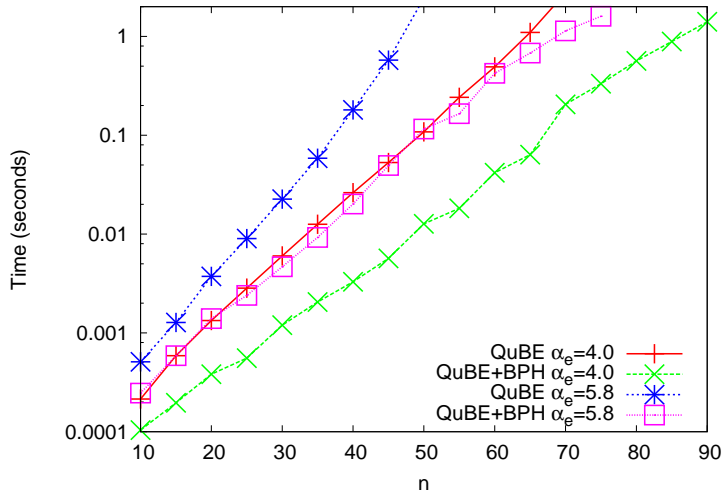}
\caption{ \label{fig:quben}
Time used by DPLL of QuBE7.2 and QuBE7.2+BPH in solving random $(1,3)$ formulas with 
fixed $\alpha_e$ values and different number of variables $n$ ($N_u=N_e=n$).
Data are averaged over $2000$ instances. As shown in Fig.~\ref{fig:50con}, with $\alpha_e=4.0$, most formulas are satisfiable 
and with $\alpha_e=5.8$, most formulas are unsatisfiable.}
\end{figure}

Our results are presented in Fig.~\ref{fig:50con}. The power of QuBE
enables to reach larger formulas than we used in
Fig.~\ref{fig:15con}, so our experiments are carried out on random $(1,3)$ formulas with $N_e=N_u=50$.
From  the figures we can see that BPH considerably reduces the size of the search
tree as well as the computation time exponentially for whole range of $\alpha_e$. 
The improvement in performance is relatively small only close to the
transition region because information given by BP is probably less
reliable there. Figure ~\ref{fig:quben} shows the computational time 
reduction with the system size. We see that with the same time limit, BPH
enables QuBE to solve larger formulas.

\begin{figure}
\includegraphics[width=0.8\textwidth]{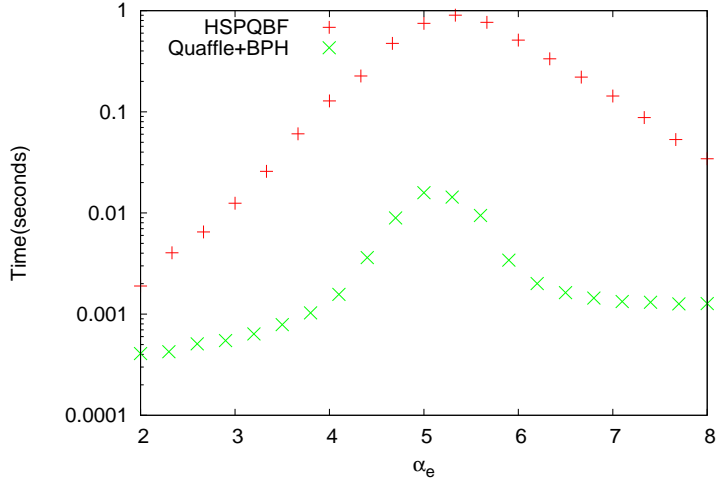}
\caption{ \label{fig:30time}
Time used by Quaffle+BPH and by HSPQBF~\cite{Yin_etal_2011} in solving random $(1,3)$ formulas, 
data are averaged over $2000$ instances.
$N_u=30$ and $N_e=30$.
}
\end{figure}

Ideas similar to ours were already explored in \cite{Yin_etal_2011},
where the authors studied an algorithm named HSPQBF that uses
Survey Propagation (SP) as decision heuristics in a QBF solver
Quaffle \cite{Zhang_Malik_QBF_2006}. We see that using BP as decision
heuristics is more stable than SP because SP has a narrow region of working
parameter. Recall that in random SAT formulas with number of variables going to infinity, 
BP reports correct marginals with constraint density ranging from zero
to the condensation transition point~\cite{KMRSZ07}.
However, SP always has trivial solution (zero messages) when the
constrain density is smaller than a value that lies relatively close 
to the SAT-UNSAT transition point~\cite{MPZ02,Mertens_etal_RS_2006}.
Moreover, hard QBF instances are often located in the region where SAT formula created by ignoring the quantifiers is
easy and SP often has trivial solution, but BP works well. 
We cross-checked these intuitions using BPH to guide Quaffle, and
compared running time of DPLL in solving random $(1,3)$ instances by BPH guided Quaffle and HSPQBF in Fig.~\ref{fig:30time}. 
The figure indicates that in the whole range of $\alpha_e$, Quaffle with BPH gives better results than HSPQBF. We also checked 
other types of random formulas, similar results are obtained.

\section{Generalization to QBF with multiple alternations}\label{sec:extention}
\begin{figure}[!th]
  \includegraphics[width=0.5\textwidth]{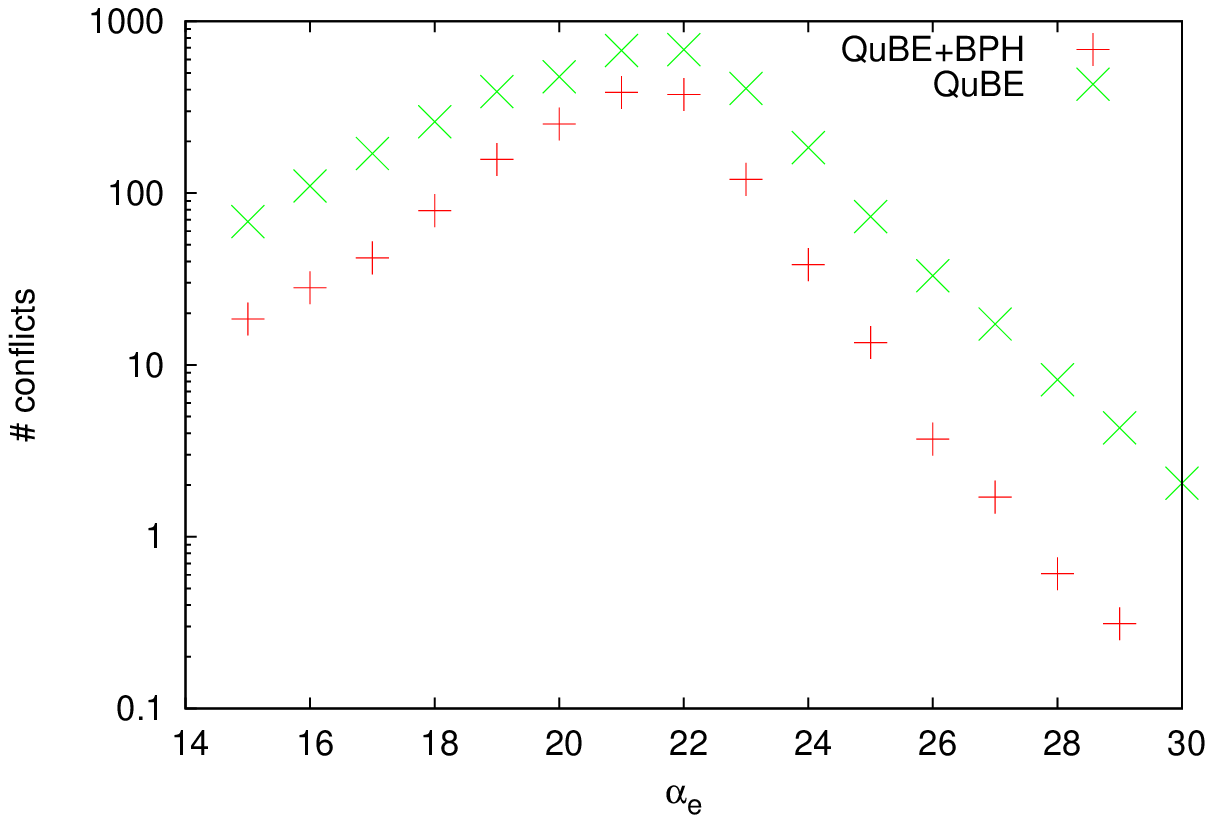}
\includegraphics[width=0.5\textwidth]{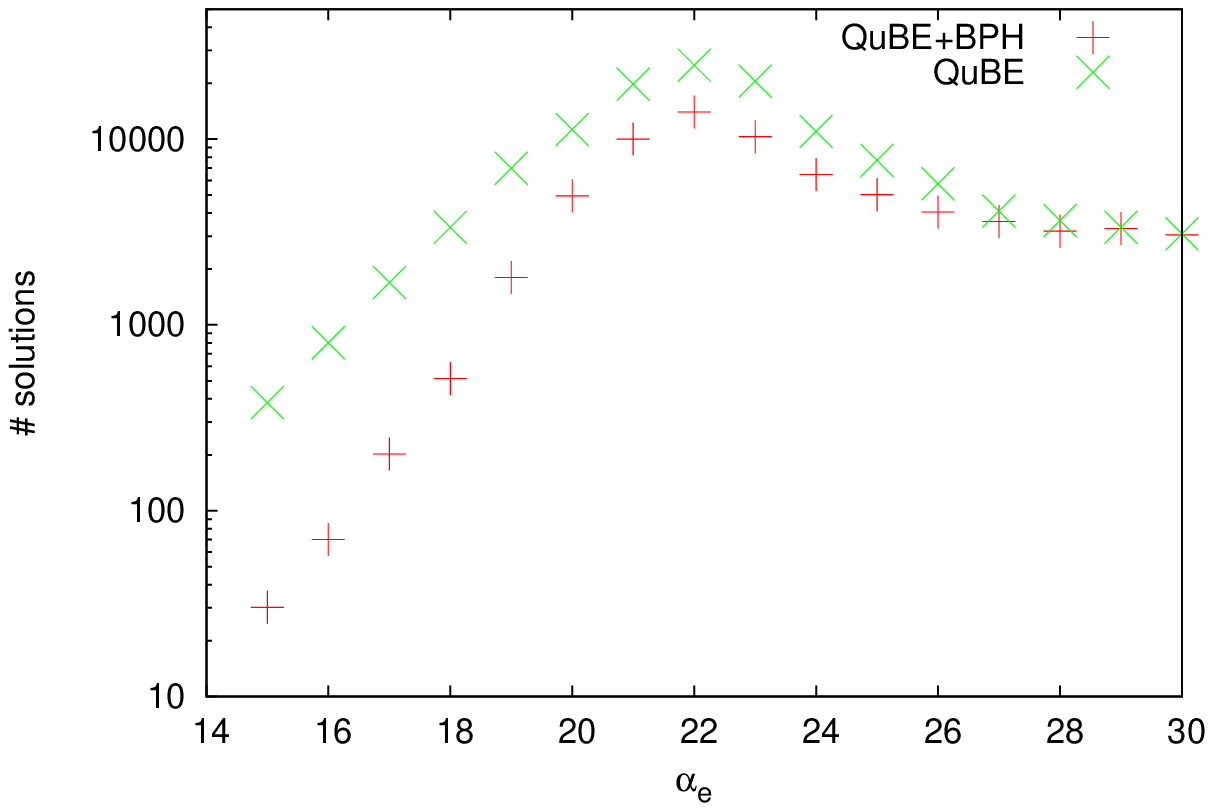}
\includegraphics[width=0.5\textwidth]{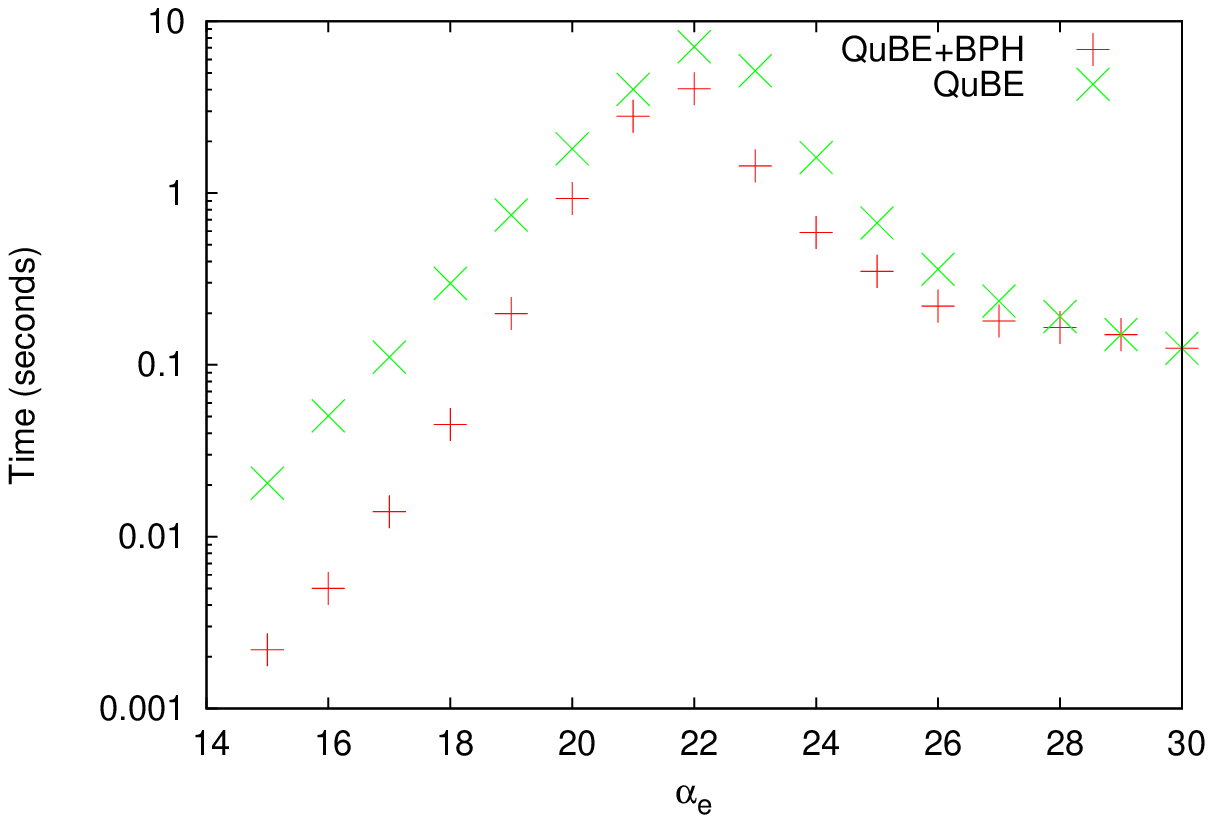}
\includegraphics[width=0.5\textwidth]{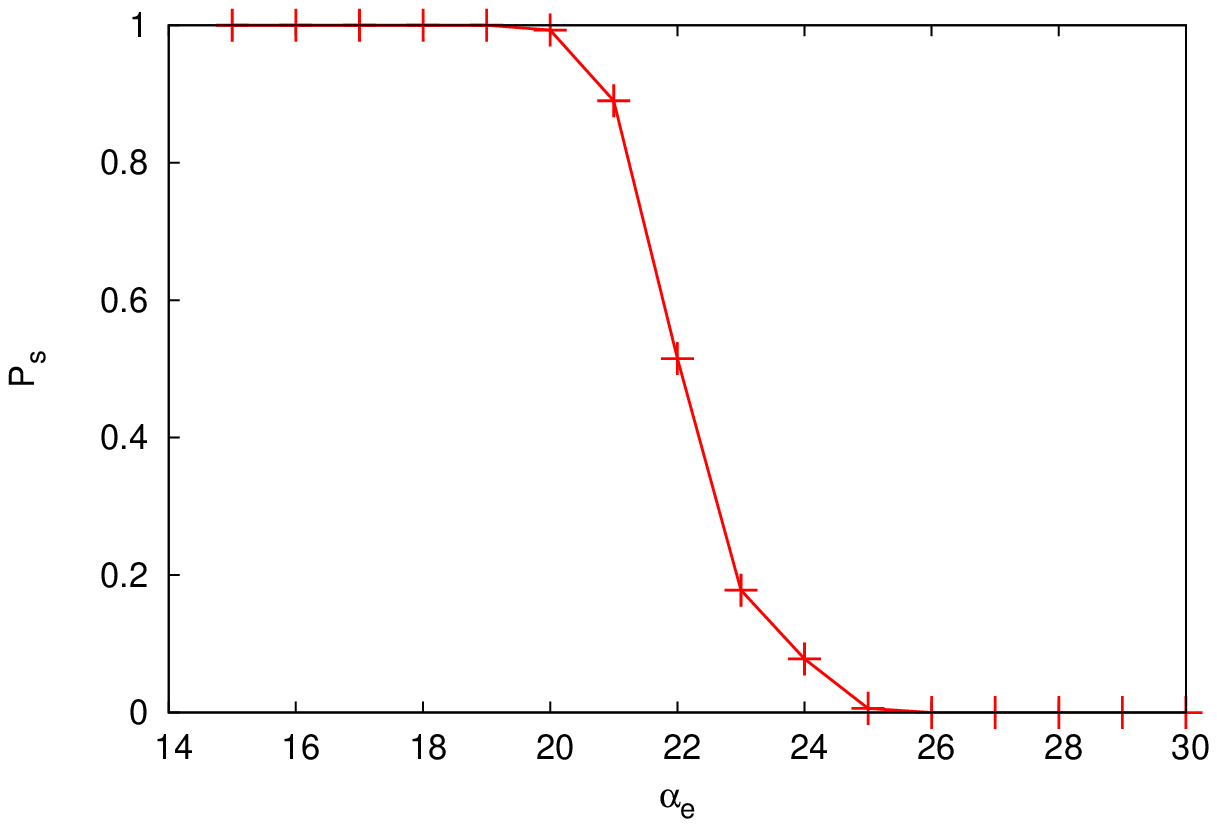}
\caption{ \label{fig:mb20con}
Number of conflicts, solutions and time used by DPLL of QuBE7.2 in solving model-B formulas, data are averaged over $200$ instances.
Formulas have $4$ alternations and $n=20$ variables in each alternation. Each clause in the formula contains 
$4$ existential variables that are selected randomly from all the
existential variables (i.e. from the 2nd and 4th level variables) 
and $1$ universal variable that is selected randomly from 
all the universal variables (i.e. from the 1st and 3rd level variables). $\alpha_e$ in figure corresponds to number of clauses divided by $n$.
In generating formula, we use formula generator downloaded from QBFLIB \cite{qbflib}.
}
\end{figure}

Our BPH decision heuristics for DPLL works naturally in QBF with
multiple alternations.
To test performance of BPH in general QBF, we tested model-B formulas with $4$ alternations, results are plotted in
Fig.~\ref{fig:mb20con}. As the figure shows, for 
small $\alpha_e$ BPH improves the performance considerably. 
However, it gives only small performance improvement for large $\alpha_e$ when all formulas are unsatisfiable.

A multi-alteration QBF can be transformed to a $2$-alternation QBF by changing the order of universal and existential variables.   
For example, given a 4-alternation QBF $\mathcal{F}=\forall X_1\exists X_2 \forall X_3 \exists X_4\Phi$, we can arrive at $\mathcal{F}'=\forall X_1X_3\exists X_2X_4$ by switching the order of $X_2$ and $X_3$. One can prove that if $\mathcal {F}'$ is unsatisfiable, then $\mathcal{F}$ is unsatisfiable.
The heuristic algorithm can be used to prove the unsatisfiability of $\mathcal {F}'$, and so that of $\mathcal{F}$. 

\section{Performance of BPH on structured formulas}\label{sec:structured}
In contrast with random formulas, BP usually does not give accurate information 
about the solution space on structured formulas because of existence of many short loops. 
Hence we do not expect BP to improve complete solvers in solving every structured formulas. 
We tested some structured benchmarks from QBFLIB~\cite{qbflib}, part of 
the results are listed in Table 1. We can see from the table that 
on some formulas, BPH increases the performance of QuBE7.2 while in other formulas, BPH decreases the performance.
Remarkably, some instances e.g. \textit{ncf\_16\_64\_8\_edau.8} problem and \textit{ii8c1-50} problem, 
which have not been solved by other solvers (in 
previous QBF Evaluations), can be solved by QubE7.2+BPH in few seconds.
 
\begin{table}[!ht]\label{table:benchmarks}
  \begin{center}
  \caption{Running time (in seconds) of QuBE7.2 and QuBE7.2+BPH in solving structured benchmarks.}
  \begin{tabular}{|c|c|c|c|c|c|c|}
	\hline
	Name of instance &   QuBE  & QuBE+BPH&   & Name of instance &   QuBE  & QuBE+BPH\\
	\hline
	ii8c1-50 & $>600$ & $14.21$& & ev-pr-8x8-15-7-0-1-2-lg & $0.023$ & $>600$\\
	\hline
	 ncf\_16\_128\_2\_edau.8& $>600$ & $0.27$ & &flipflop-12-c  & $0.71$ & $>600$\\
	\hline
	 ncf\_16\_64\_8\_euad.4& $>600$ & $4.01 $ & &lut4\_3\_fAND  & $ 0.25$ & $>600$\\
	\hline
x170.5 &$ 520.44 $&$ 63.42 $&               &                cf\_2\_9x9\_w\_ &$ 0.18 $&$ 539.49 $\\
	\hline
ncf\_8\_32\_4\_euad.4 &$ 301.86 $&$ 0.14 $&              &      ncf\_4\_32\_4\_u.9 &$ 35.69 $&$ 565.35 $           \\
	\hline
szymanski-5-s &$ 153.00 $&$ 19.31 $&              &                 c2\_BMC\_p2\_k8 &$ 95.49 $&$ 207.25$\\
	\hline
k\_d4\_n-5 &$ 408.15 $&$ 83.94 $&             &                 toilet\_a\_10\_01.15 &$ 7.45 $&$ 510.84 $\\
	\hline
ncf\_4\_32\_8\_edau.3 &$ 349.78 $&$ 0.25 $&    &                 cf\_3\_9x9\_d\_ &$ 0.17 $&$ 480.27 $\\
	\hline
x170.19 &$ 340.05 $&$ 13.07 $&               &                 connect\_9x8\_8\_D &$ 0.08 $&$ 115.79$\\
	\hline
k\_grz\_p-13 &$ 190.66 $&$ 24.99 $&    &                stmt21\_252\_267 &$ 3.93 $&$ 142.49 $ \\
	\hline
connect\_5x4\_3\_R &$ 97.51 $&$ 38.35 $&     &                k\_ph\_n-20 &$ 7.65 $&$ 462.44 $  \\
	\hline
 BLOCKS4i.6.4 &$ 49.21 $&$ 29.87 $&         &                 toilet\_a\_10\_01.11 &$ 0.02 $&$ 416.16 $ \\
	\hline
C880.blif\_0.10\_1.00& & & & & &  \\

\_0\_1\_out\_exact &$ 93.21 $&$ 44.45 $&     &x115.4 &$ 201.46 $&$ 302.64 $\\
	\hline
CHAIN18v.19 &$ 346.39 $&$ 288.12 $& &stmt21\_143\_314 &$ 92.61 $&$ 92.75 $\\
	\hline
  \end{tabular}
  \end{center}
\end{table}

\section{Conclusion and discussion}

In this paper we developed heuristic and complete algorithms for QBF based on message passing. Our heuristic 
algorithm BPDU and BPSPDU use message passing to find a universal assignment that evaluates to an unsatisfiable remaining formula, 
thus to prove the unsatisfiability of $2$-alternation QBF. Our complete algorithm is based on DPLL process, 
which searches the whole configurational space by backtracking more efficiently using message passing branching heuristics 
BPDH and BPH. Our algorithms described above can be downloaded from \cite{algorithms}.
Both our algorithms work very well on random QBF and in some cases they provide large improvement also on structured QBF and 
solve some previously unsolved benchmarks in QBFLIB. These results
should encourage further investigation of the use of message passing
as heuristic solvers or as guides for heuristics included in DPLL-like
complete solvers.

\section{Acknowledgements}
R.Z. and P.Z. would like to thank Massimo Narizzano for discussing and
sharing the source code of QuBE7.2. 
P.Z. would like to thank Minghao Yin and Junping Zhou for discussing
and sharing the source code of HSPQBF.


\end{document}